\documentclass[review]{elsarticle}

\usepackage{lineno,hyperref}
\biboptions{authoryear}

\begin{document}

\begin{frontmatter}

\title{The Mode of Computing}

\author{Luis A. Pineda\fnref{myfootnote}}
 \address{Universidad Nacional Aut\'onoma de M\'exico}
\fntext[myfootnote]{lpineda@unam.mx}




\begin{abstract}

The Turing Machine is the paradigmatic case of computing machines, but there are others such as analogical, connectionist, quantum and diverse forms of unconventional computing, each based on a particular intuition of the phenomenon of computing. This variety can be captured in terms of system levels, re-interpreting and generalizing Newell's hierarchy, which includes the knowledge level at the top and the symbol level immediately below it. In this re-interpretation the knowledge level consists of human knowledge and the symbol level is generalized into a new level that here is called \emph{The Mode of Computing}. Mental processes performed by natural brains are often thought of informally as computing process and that the brain is alike to computing machinery. However, if natural computing does exist it should be characterized on its own. A proposal to such an effect is that natural computing appeared when interpretations were first made by biological entities, so natural computing and interpreting are two aspects of the same phenomenon, or that consciousness and experience are the manifestations of computing/interpreting. By analogy with computing machinery, there must be a system level at the top of the neural circuitry and directly below the knowledge level that is named here \emph{The mode of Natural Computing}. If it turns out that such putative object does not exist the proposition that the mind is a computing process should be dropped; but characterizing it would come with solving the hard problem of consciousness.

\end{abstract}


\end{frontmatter}

\newtheorem{mydef}{Definition}
%\linenumbers

\section{Cognition and Computation}

Cognition is the study of mental processes, such as perception, thought, attention and memory. It has been with us at least since the Greeks, certainly since Aristotle --e.g., \citep{Shields-on-aristotle}-- but in current times it is understood as the study of the mind in terms of information processes or, more specifically, as computational processes. This use was anticipated by Charles Babbage's Analytic Computing Engine (ACE) in the XIX century, that could automate the computation of mathematical functions, and the manipulation of symbols more generally, but it was properly introduced with the Turing Machine (TM) in 1936 \citep{TuringMachine} when an abstract and general notion of computing was presented. The TM allowed to model mental processes, such as playing chess, a paradigmatic form of thinking, that could only be performed by people before computers were available. The potential scope of computing machinery for simulating the mind was stated explicitly by Turing with the presentation of the Imitation Game, better known as the Turing Test, and with his expectation that digital computers will eventually compete with humans in all purely intellectual fields \citep{Turing}. Turing also held that a computing engine able to pass the imitation game should be ascribed intentionality and consciousness, in the same way that we ascribe such properties to other people, regardless that we only have access to our own internal subjective experience. This position gave rise to the more general claim that that the mind is a computational process carried on by a TM. This is the central tenet of current cognitive studies. This view holds implicitly that the TM, which is a human invention, captures a putative phenomenon which can be designated as \emph{natural computing}, that was never observed before the invention of computers.

From a mathematical perspective, TMs compute functions --see \citep{Boolos-Jeffrey}. Computing is simply the process of obtaining the object in the codomain --i.e., the value-- assigned to a given object in the domain --i.e., the argument-- for all the objects in the function's domain. This process is achieved by means of an algorithm that is executed by the TM. If the algorithm terminates for all of its arguments is said to be effective, and the functions that have an effective algorithm are said to be computable. Functions that do not have an effective algorithm, or have no algorithm at all, are said to be uncomputable. TMs are enumerable and can be placed in a list including all possible algorithms. The hypothesis that all computable functions are computed by one or another algorithm in such a list is called Church's Thesis or Church-Turing Thesis. Church's Thesis also states that the TM is the more general computing engine that can ever be conceived, and that every alternative general enough computing device computes the set of functions that are computed by the TM exactly. The Thesis postulates or defines the very notion or concept of computability, and is the fundamental tenet of computer science. However, the inverse problem of finding out the algorithm that computes a given function, for all computable functions, is open --it would be required to know in advance whether the function is computable or not by means of a computing engine more powerful or qualitatively different in an essential way from the TM, which cannot exists by hypothesis, and the Thesis cannot be proven. Conversely, the Thesis would be refuted if such a machine were discovered.

Adopting the current view of cognition formally --as opposed to metaphorically-- involves postulating that the mind is a computing process, that Church Thesis is true, and that natural computing is powered by the TM. Consequently, that the mind is produced by a set of effective algorithms and cannot compute uncomputable functions. However, it might turn out that there is natural computing but powered by a machine that is not equivalent to the TM in an essential way, so the mind could compute uncomputable functions in Turing's sense; hence, Church Thesis would be refuted or at least its scope would be limited; and it is also possible that the mind is not caused by a computing engine and there is not natural computing after all. 

\section{Computation, Representation and Interpretation}

Computing machines manipulate representations but do not make interpretations. This is shown very clearly in computing engines built with mechanical devices. Babbage's Analytical Engine, for instance, used mechanical gears to express and perform computations \citep{Rojas-2021}. The input argument and the output value of the computed function were represented by the gears positions when the computation was started and ended, respectively, in the same way that mechanical clocks display the time. The computing process was performed by the mechanical machinery that ran continuously until it came to a halt, but the machine did not have the functional nor the structural ingredients to interpret such representations --did not understand or had knowledge in any conceivable sense-- and the interpretation was performed by people, in the same way that the positions of the clock hands are interpreted as the time by humans, but clocks do not understand the concept of time.

Representation are material objects that use a physical medium with patterns marked on it that express information. Intuitively, natural representations appeared when the phenomenon of communication first emerged, most probably very early in the story of life, perhaps as movements or gestures to express basic intentions such as where to go to find food or shelter, or run away from a predator. The representations are \emph{produced} and \emph{interpreted} by intentional agents. The former is a process that ``places'' a mental content or intent on a physical medium, and the latter is the reverse process that inputs the representation and places its content in the interpreter's mind. Hence, the production and interpretation of representations indicates clearly the the agent has a mind already.

Producing and interpreting representations are subsumed within the more general processes through which intentional agents perform actions with intent and interpret the material world. Such actions and interpretations depend on the particular brains, that support the mental endowment of natural species and individuals, such as perception and motor behavior and, in the case of humans, thought. Hence, the mental world is constituted by interpretations, that are subjective, as opposed to the objective material reality. Consequently, representations, that express mental content, represent interpretations.

A representation is \emph{declarative} if the result of interpreting it places the concept of an individual object, property or relation in the interpreter's mind. For instance, the speech act ``this is a ball'' accompanied by a pointing gesture, is a declarative representation. There are also representations that express instructions or commands that the interpreter should perform directly. In this case, interpreting the representation consists on performing an action, such as collecting food or running away from a predator. This kind of representations are \emph{procedural}. Performing actions is accompanied by an experience or feeling, and interpreting the representation and experiencing performing the actions, are two aspects of the same act. Representations can have a dual declarative and procedural aspect, as the interpretation can produce mental content, that the interpreter may be conscious of, and involve performing actions, that the interpreter may experience or feel.

The paradigmatic form of human natural representation is the spoken language. The medium is the air wave, and the sound patterns codify the phones, words and sentences. Speech acts, either spoken, gestural or multimodal, express the knowledge, beliefs, desires, intentions, feelings and affections of the agent. Natural representations evolved into conventional representations that were not necessarily linguistic, such as diagrams, paintings and sculptures, with deep and subtle meanings, that allowed the transmission of mental contents beyond the immediate face to face interaction. Then came the invention of manuscript symbols and the appearance of written language, that uses the paper as the physical medium and the ink to mark the patterns, and allowed communication between agents at different spatial and time locations. The written language served also as the first form of artificial memory, that allowed registering commercial, biographical or historical records, and the appearance of literary art. It also made possible, with the invention of numerical notations and algorithms, the ability of making complex calculations beyond what is possible to calculate mentally. The next great chapter of the history of representations was the invention of the printing press. Gutenberg automated the production of textual representations using movable printing types with standard forms for letters, digits and punctuation marks, that allowed the impression of different texts through a manual but regimented process, using a finite number of types, making possible the massive production of texts, with the consequent cultural explosion.

The Turing Machine can be thought of as the next main cultural evolutionary event in the history of representations. The invention was made in the tradition of Babbage's Analytical Engine, which was already a universal computer, but Turing introduced the printed text as the representational format, instead of Babbage's gears. While Gutenberg's machine automated the process of printing texts, the TM automated their transformation through mechanical well-defined procedures. The TM uses a tape divided into cells as its representational medium, each holding an instance of a type symbol of a given finite alphabet, and the tape as a whole holds a textual representation. The machine can be in one of a finite number of possible states, and has a scanning device placed on a specific cell at the current state. It also has a control system that executes the instructions included in the so-called transition table. Every instruction specifies the action that is taken at each state and symbol placed on the current scanned cell, and the state that the machine goes once the instruction is performed. The actions that the machine can do are only to write a symbol on the cell being scanned, or to shift the scanner one cell to the right or to the left. The machine halts when the action and next state for the current state and symbol being scanned is not specified in the transition table. The instructions specify the algorithm, and all computable functions are computed by this very simple machine. 

The interpretation of the textual representations on the machine's tape is performed by people in relation to a well-defined set of interpretation conventions, as stressed by Boolos and Jeffrey \citep{Boolos-Jeffrey}, but not by the TM that lacks the functional and structural ingredients to do so. A computing engine involves these two aspects always: the device or natural phenomenon that maps or transforms the inputs into the outputs, and the agent that makes the interpretation or the semantic attribution. Put it simply and directly, there is no computation without representation and interpretation.

Textual representations where later ``internalized'' into digital computer with the development of the von Neumann architecture and the introduction of the Random Access Memory (RAM). In these machines each basic memory cell or byte contains an instance of a type --e.g., a symbol of the ascii alphabet-- representing a letter, a digit or a punctuation mark, and the content of a physical or logical sequence of bytes is a textual representation. Algorithms are expressed through computer programs as strings of texts, that are compiled and placed into the computer memory as binary codes, which are also textual representation using only two types of digits. Images of the diverse modalities of perception and action are codified as strings of digits too. Modern digital computers are implementations of the TM that manipulate representations, but the interpretations are made by people, and computer machines know, experience and feel as much as carving tools, pens and the printing press.

Written text, specially in the printed form, is perhaps the paradigmatic form of representation, and its adoption as the representational format by Turing contributed greatly to the success of the TM and to the impact of computing theory and technology. However, this choice prevents the direct use of non-textual representations, natural or conventional, such as hand and body gestures, diagrams, pictures, paintings, sculptures, music pieces, etc., that need to be translated into a textual format, using digits and numerical notations, to be used by computers. Although science and technology have addressed this challenge very successfully, this interface process is nevertheless complex, and it is unlikely that has to be performed by people. 

This limitation impacts also in representing the natural world that is presented to the agent through signals and forces, giving rise to images of the different modalities of perception and action, that have to be translated into textual representations in order to be used by computers. This is also a huge technological challenge, and despite that it has been addressed very successfully too, it is nevertheless a strong interface problem, that is unlikely to be faced by humans and non-human animals.

The objective character of representations and its use in digital computers contrasts with the subjective notion of representation in cognitive psychology and philosophy of mind. Informally, representations in these latter disciplines are mental objects that are causal and essential to thought, intentional behavior and consciousness. Although these putative mind objects cannot be observed, measured or characterized, they were nevertheless used in pre-scientific speculative or introspective methods, but later on were rejected as genuine objects of scientific study, as in behaviorism. However, the TM and the availability of practical digital computers suggested the use of TM representations for modeling natural cognition. Chomsky made the proposal explicit with the introduction of syntactic structures \citep{Chomsky-1957} and his refutation of Skinner's \emph{Verbal Behavior} \citep{Chomsky-1959}. Chomsky's implicit move was to establish a direct analogy between the material ``external'' representations and the representations processed by the mind and held in memory, which are ``internal''. More generally, the representational view of cognition was extensively argued by Fodor \citep{Fodor} who proposed that the object of cognition was a mental language, the Language of Thought (LOT) or \emph{Mentalese}, with a syntactic or compositional structure and a propositional and representational character, that was inspired in the TM directly.

However, the simple internalization of textual representations into digital computers does not make them intentional, and the TM cannot be the engine that powers natural computing. Hence, in the absence of an alternative to Turing's notion of computing, the whole idea of natural computing should be dropped.

\section{Knowledge and System Levels}

\begin{figure}
\includegraphics[width=0.5\textwidth]{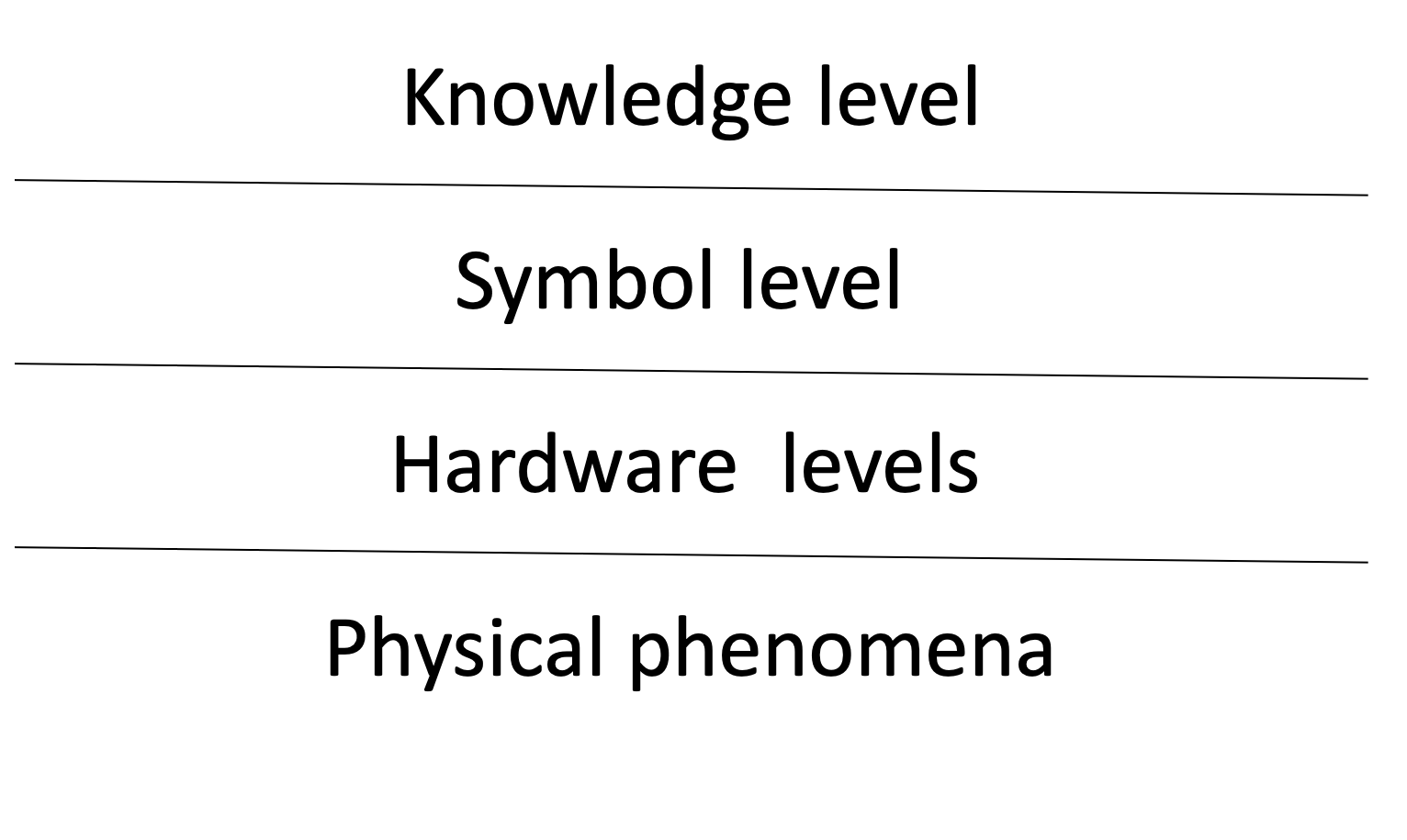}
\centering
\caption{Newell's main system levels}
\label{System-Levels}
\end{figure}

The distinction of the two senses of representation --material versus mental-- can be seen in terms of system levels. Complex phenomena or devices can be studied at different levels of abstraction or granularity, such that each level has its own theoretical terms and laws of behavior, and the phenomena at each level can be studied independently of other levels. For the case of digital computers Allen Newell distinguished, from bottom to top, the physical phenomena, the device, the electronic circuits, the logic circuits, the register transfer or computer architecture levels, with the symbol and the knowledge levels at the top of the hierarchy \citep{Newell}. A simplified version of Newell's levels collecting the hardware levels together is illustrated in Figure \ref{System-Levels}.

The TM proper is defined at the symbol level, where symbolic manipulation takes place, and reduces to all the levels below, down to the physical phenomena. Meaning, on its part, belongs to the knowledge level. Newell stated that the knowledge level emerges from but is not reducible to the symbol level.  He also sustained that the medium of this level is knowledge itself and that the only rule of behavior at this system level is what he called the \emph{Principle of Rationality}. He --in conjunction with  Simon-- also postulated the  \emph{Physical Symbol System Hypothesis} sustaining that a system of grounded symbols provides the necessary and sufficient condition to generate general intelligence \citep{Newell-Simon}. This latter hypothesis refers also to the symbolic manipulation performed at the symbol level. From Newell and Simon's claims it can be elicited that the knowledge level holds the representations that result from interpreting the structures at the symbol level, thus digital computers can understand and be conscious.

According to the previous discussion this latter claim cannot be sustained, but Newell's hierarchy can be still adopted, but holding that the knowledge level contains the interpretations ascribed to symbolic structures by people. Hence, in the present re-interpretation of Newell's hierarchy, the symbol level and all the levels below, down to the physical phenomena, are implemented in machines, but the knowledge level is human knowledge.

An alternative system level hierarchy was proposed by Marr \citep{Marr}. This has three levels which are, from top to bottom, the computational or functional, the algorithmic and the implementational. The first refers to the specification of the mathematical function that models the mind's faculty that is the object of study, such as vision or language; this is human knowledge and the top level in Marr's hierarchy corresponds to the knowledge level. The algorithmic level is constituted by the computer programs or algorithms that compute the function properly and correspond to the symbol level. Finally, the bottom level includes all hardware and software aspects that sustain the algorithmic level, but are contingent to the particular computations, such as the programming language in which the algorithm is coded, or the particular computer in which it is run.

From a third perspective, the sense of representation at the knowledge level corresponds to the one contested by Searle in the story of the Chinese Room, in which an English speaking person, who does not understand Chinese, answers nevertheless questions in Chinese by following instructions and data, without been aware of the meaning of the questions and their answers \citep{Searle}. In terms of the hierarchy of systems levels, all operations of the person in the room are performed at the symbol or algorithmic level, but such individual is not aware or conscious of the meaning of the Chinese expressions because he or she does not represent at the knowledge level. Searle dubbed the view that computers represent at the symbol level but do not do so at the knowledge level \emph{weak AI}, in opposition to the view that computers can represent at the knowledge level, that Searle refers to as \emph{strong AI}.

The attribution of meaning to computing systems should also be seen in relation to whether subjects of study are actual mechanisms, such as robots, or whether computers are used for developing and testing cognitive models. In the former, and considering that TMs do not ascribe meaning, such devices are similar to standard machinery --i.e., radios, TVs, etc.-- which are unaware of the meanings of the signals they manipulate for human consumption. In the latter, computers are unaware of the meaning of the representations that they are supposed to model too, but the theorists who devise such models interpret and attribute meanings to such processes. So, once again, computer systems implement cognitive models at the symbol level but interpretation and representation belong to the knowledge level which resides in the mind of human experimenters.

Newell's, Marr's and Searle's models refer to standard computations using the TM, but do not intent to address the case of the putative natural computing. However, these views can be used as analogies of how the mind could relate to the brain: there would be a natural system level implemented in the brain, corresponding to the symbol level, that would specify the actual codes through which information is represented in the brain and the operations performed upon them. Such natural computing level would be directly below the knowledge level --the level of interpretations-- and at the top of the brain's system levels.

\section{Connectionism and alternative notions of Computing}
\label{connectionism}

The view that the mind reduces to a simple symbol manipulation machine for making arithmetic calculations or linguistic inferences has been questioned, and there are alternative intuitions of what is computing, especially natural computing. On such proposal is the Parallel Distributed Processing (PDP) program or Connectionism and its implementation through Artificial Neural Networks (ANNs). This program holds that the mind is a computational process too, but that intelligence emerges from the interactions of large numbers of simple processing units expressing \emph{distributed representations}, and that symbolic systems or TMs have failed to model appropriately most mind processes, such as perception, memory, language and thought --see the Preface of Rumelhart's text \citep{Rumelhart}. Distributed representations are opposed to TMs in the relation between units of memory and the units of content or the concepts that such structures represent: in TMs this relation is one-to-one, while in distributed systems is many-to-many \citep{Hinton-1986}. Thus, while the basic concepts are represented in mutually exclusive memory regions in the TM, the concepts expressed by distributed representation use memory regions that may overlap in arbitrary ways; for instance, a particular neuron can contribute to the representation of more than one concept, and a concept may share neurons with other concepts. In addition, distributed representations can generalize and the extension of the represented concept may contain individuals that were not considered when the representation was originally formed if they are similar enough to other individuals in such extension. This latter effect cannot be expected in local representations.

\begin{figure}
\includegraphics[width=0.5\textwidth]{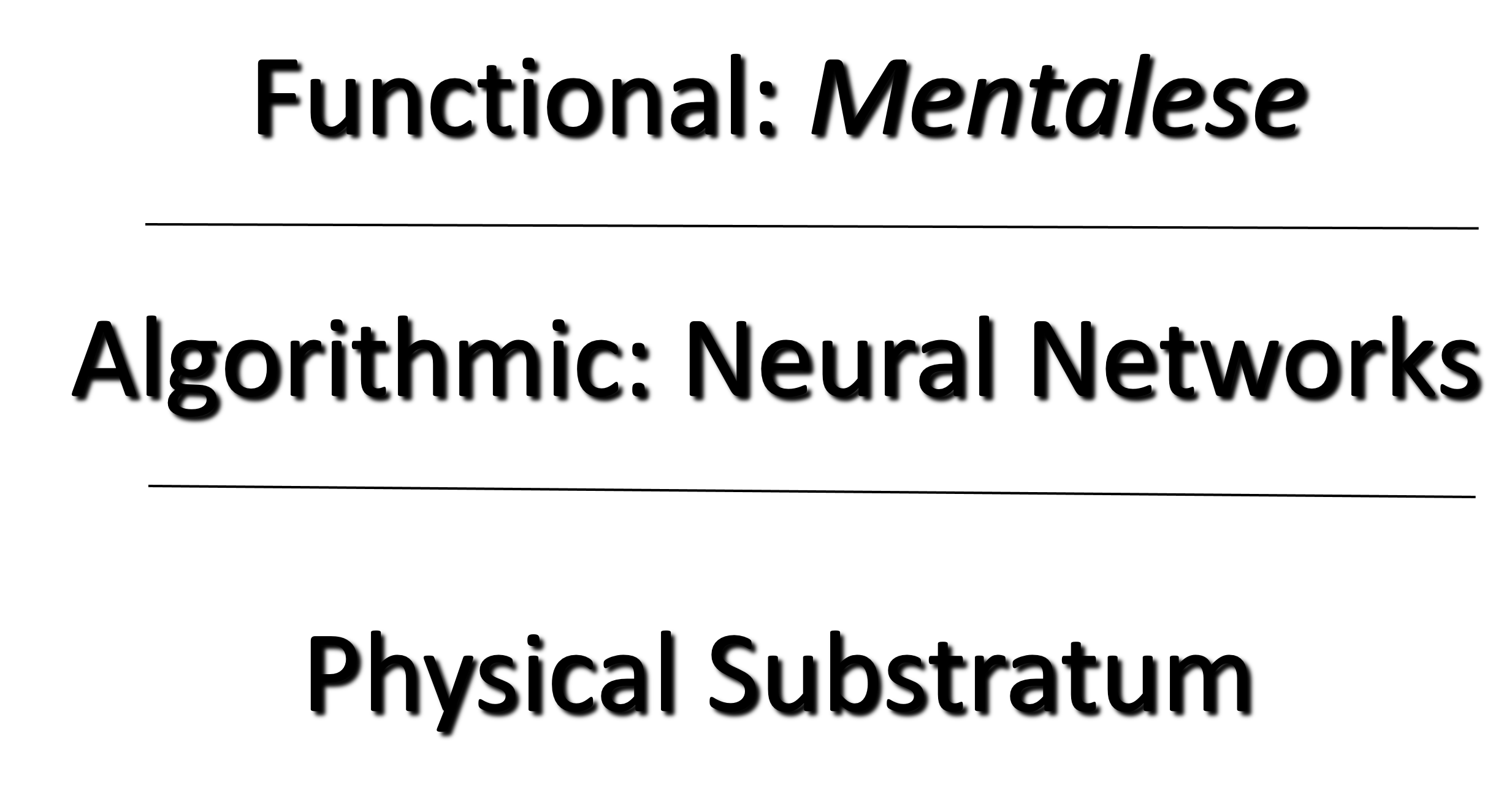}
\centering
\caption{Mentalese and its implementation in ANNs}
\label{Fodor-Architecture}
\end{figure}

Whether symbolic and connectionist systems are different models of computing has been subject of intense debate. Fodor and Pylyshyn, for instance, emphasized that both symbolic and distributed systems are meant as representational, but highlighted the limitations of ANNs to express syntactic structure and for holding information as standard declarative memories, and denied the special character of connectionist algorithms \citep{Fodor-Pylyshyn}. Nevertheless, they granted that connectionist systems could implement the mentalese, which emerges from the workings of the natural neural networks. This relation has been described in terms of system levels \cite{Rueckl}, but using Marr's system levels \citep{Marr}, as illustrated in Figure \ref{Fodor-Architecture}. 

A very large effort has been made to reduce the mentalese to connectionist systems; in the computational front this is interpreted as expressing symbolic representations and reasoning through ANNs \citep{Chalmers, Lecun, Lecun-nature} or as integrating ANNs, Bayesian and logical systems \citep{Besold-2017}; and implementing symbolic representation and reasoning with distributed representations is possibly the greatest challenge for current Deep-Learning Artificial Neural Networks (DL-ANNs) and there are large efforts to achieve this end --e.g., \citep{Graves-2016}. 
 
However, connectionism and ANNs deviated from the original program in practice because ANNs are specified as Turing-computable functions, implemented with standard data-structures and algorithms, and run in standard digital computers, thus ANNs are TMs. ANNs are expressed and computed with numerical matrices and  operations, alike in every respect to standard scientific or engineering applications. It has also been shown that every TM can be expressed as a recurrent neural-network and these are equivalent formalisms in relation of the set of functions that can compute \citep{Siegelmann,Sun}. Hence, implementing ANNs with TMs fuse the symbolic and the connectionist at the algorithmic system level, and the interpretation is performed according to the standard configuration and interpretation conventions of the TM. If the claim is that ANNs express distributed representation, but practical implementations use local memories, the \emph{distributed property} is only simulated but not actual. Furthermore, the global interpretation of the distributed representations is dissolved and is only present as a subjective intuition of the human interpreter. Hence, ANNs, including current DL-ANNs, contrary to Rumelhart's claim, do not contest the TM, and do not challenge Church Thesis.

\section{The Computational Quadrants}
\label{quadrants}

The distinction between declarative and procedural representations shows up in Artificial Intelligence (AI) as the strong opposition between \emph{symbolic} versus \emph{sub-symbolic} systems. According to the so-called Knowledge Representation Hypothesis, the knowledge exhibited by a computational process is considered symbolic or ``representational'' if its structural ingredients can be rendered as propositions with a linguistic character, that are also causal and essential to the behavior exhibited by the computational agent \citep{Brian-Smith}; otherwise, the systems are said to be sub-symbolic or ``non-representational''. The former kinds of systems are declarative whereas the latter have implicit knowledge embedded in opaque structures --e.g., numerical algorithms, neural networks, etc.-- which is used or deployed, and are procedural. However, the Knowledge Representation Hypothesis does not commit to whether or not computers make the semantic attribution.

Strong and weak AI are symbolic, but while the former is meant as representational, in the sense that such systems make interpretations, can be conscious and can experience the world, the latter is understood as non-representational, so computers carry on with the symbolic manipulation but people make the interpretation. ANNs and diverse algorithmic approaches are sub-symbolic, but are meant as representational too, at least in its initial formulation, and are alike to strong AI in this respect. There are also sub-symbolic approaches that are meant as non-representational. This seems to be the position adopted in Brooks's program of the so-called embedded architectures for modeling bio-inspired robotic mechanisms \citep{Brooks,Brooks-IJCAI-1991}. Brooks's move was to construct robots with sensors and actuators of different sorts, including control and dynamical systems implemented physically or simulated with digital computers using procedural representations; however, the computing engine was no longer meant to be causal and essential to the behavior exhibited by the robot, the basic tenet of AI, and was only used as the modeling tool, as in most current scientific disciplines. Brooks dubbed his view with the lemmas ``Intelligence without Representation'' and ``Intelligence without Reason'' challenging directly the cognitive and AI tradition. Nevertheless his program gave rise to a large body of work developed within the so-called \emph{Embodied Cognition} and is considered an AI paradigm \citep{Anderson-b:2003}. In any case, Brooks's robots do not make interpretations by hypothesis, and the view is alike to Searle's weak AI in this respect. Brooks's proposal can be included in a psychological current of thought that goes back to functionalism, classical behaviorism, Gibson's ecological psychology and situated cognition, including enactivism \citep{Froese}, which is non-representational and non-computational, and opposes the computational view of mind and AI \citep{Chemero-2013}.

\begin{figure}
\includegraphics[width=0.7\textwidth]{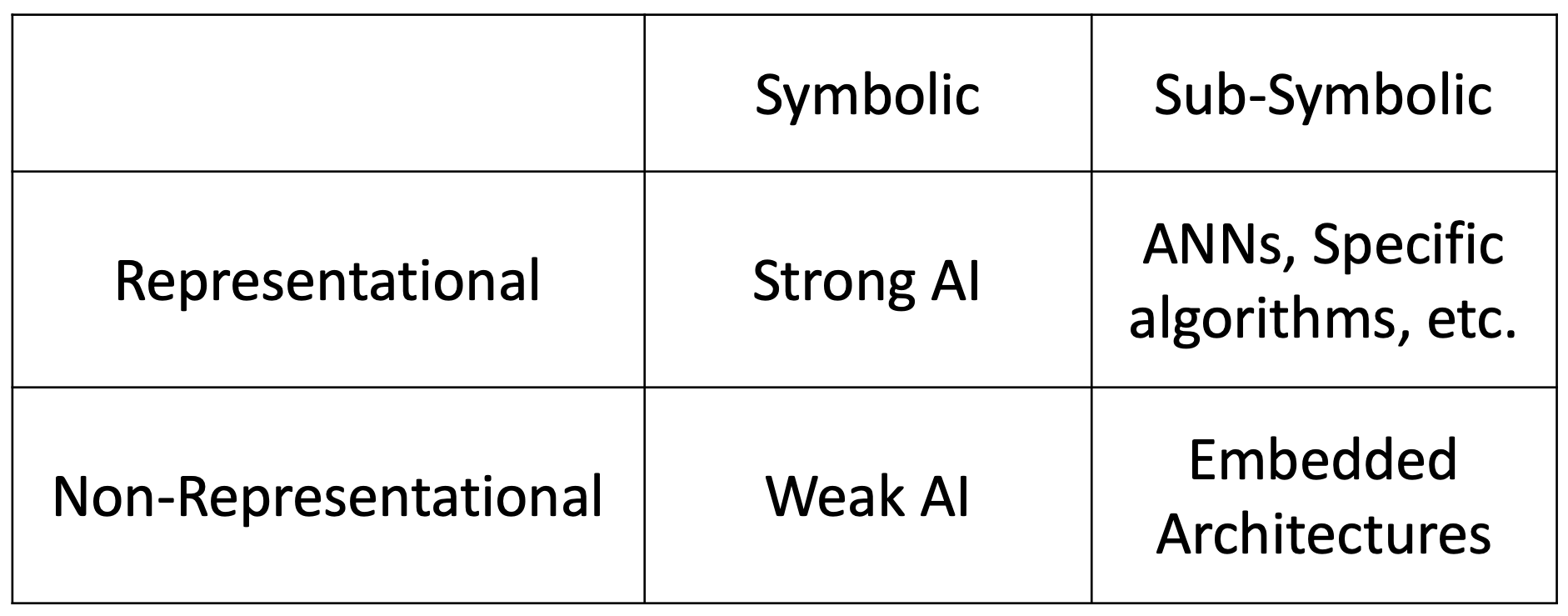}
\centering
\caption{Computational Quadrants}
\label{Comp-Quadrants}
\end{figure}

In summary, AI systems are either symbolic or sub-symbolic, according to whether they are declarative versus procedural, and representational versus non-representational, depending on whether or not the machine is assumed to make the semantic attribution and/or experience the world. The AI views in the four quadrants resulting from the combination of these two variables with their corresponding values is illustrated in Figure \ref{Comp-Quadrants}. All four quadrants are implemented with the standard digital computers, hence use local representations, and are interpreted according to the standard configuration and interpretation conventions of the TM.

The four quadrants are seen from the perspective of the human interpreter as follows:

\begin{enumerate}
    \item Top-Right: the human makes conscious interpretations and thinks that the computer does so as well;
    \item Bottom-Right: the human makes conscious interpretations but thinks that the computer does not do so;
    \item Top-Left: the human interpreter thinks that the computer experiences the world;
    \item Bottom-Left: the human interpreter thinks that the computer does not experience the world;
\end{enumerate}

Of course, being conscious and experiencing the world are not mutually exclusive mental behaviors, and humans may have feelings associated to interpreting declarative representations, and experiencing the world may be accompanied to making conscious interpretations. Likewise, robots using symbolic representations for thinking and talking in addition to sub-symbolic representations supporting perception and motor behavior, would be thought of as having conscious content and experience the world in the strong AI view; but in the weak AI view, robots would be thought of as lacking consciousness and experience, even if their behavior is un-distinguishable from the behavior of humans and non-human animals in every respect.

Artificial Intelligence has today a very visible position in science and technology, and has created great social expectations. DL-ANNs, reinforcement learning and machine learning in general, have permitted the creation of very sophisticated classification, prediction and diagnosis techniques that have allowed the development of very sophisticated natural language and vision systems, as well as a large number of scientific applications, with great actual and potential impact. This trend will continue most probably for the foreseeable future. However, the ``mental competence'' of today's most sophisticated programs and robots is very limited when is compared with humans and non-human animals with a developed enough nervous system, and it is possible that there are limits to the explanatory power of the computational view of the mind. The challenge is not to AI but most fundamentally to the whole notion of cognition based on the TM. Hence, the dilemma is whether to abandon the computing paradigm and use computers only as modeling tools, or extending the TM to model better the phenomena of the mind.

\section{Relational-Indeterminate Computing}
\label{Relational-Computing}

Turing Machine computations are determinate by necessity. Turing stated that the determinism of the machine is ``rather nearer to practicability than that considered by Laplace'' (Turing, 1950, s.5). This corresponds to the popular intuition that if something is programmed it is predetermined. From this perspective, what an algorithm does is to render explicitly the knowledge that is implicit in the definition or description of the function that it computes. Thus, if an algorithm terminates it always produces the same value for the given argument. In the original 1936 paper Turing distinguished between ``the automatic'' and ``the choice'' machines; the corresponding transition tables specify a function and a relation, respectively, so that the former machine has only one possible action at every state at the most, while the latter can have many. This kind of indeterminacy shows up in the interpretation of ambiguous sentences or in the exploration of a problem space in which more than one solution can be found for a given problem, such as equally good moves in a chess position. However, all relations can be defined as compositions of functions, and computing a relation is construed as computing its constituent functions, although at the cost that its intrinsic indeterminacy and information content is not considered. Turing did not use choice machines further in the 1936 paper, but they were latter developed in the theory of non-deterministic automata e.g., \citep{Hopcroft}, which was well established when the 1950 paper was published. Hence Turing must had been well aware of this kind of non-determinism when he compared the determinism of digital computers with the one advocated by Laplace.

Turing himself suggested an stochastic extension of the machine in the same paper, where he proposed to use the next digit of the expansion of $\pi$ as a random element. However, although such number is not known in advance, it is predetermined, and computations using such pseudo-random numbers are still fully determinate. Furthermore, he stated that it would not be normally possible to distinguish whether such machine involve genuine or pseudo-random elements. Nevertheless, he anticipated stochastic algorithms and suggested two problem solving strategies, which he called the systematic and the learning methods. The former is deterministic but the latter uses a random number and is stochastic, which Turing suggested may be more efficient in large problems that may have many solutions. The observation gave rise to stochastic computing that employed pseudo-random numbers, and later on random numbers produced out of natural signals that could be measured in the environment, and whose value could not be predicted. However, the random parameter can be thought of as a hidden or unknown argument of the function being computed, and such form of stochastic computing is still subsumed within the TM paradigm. Stochastic algorithms are programmed and executed with standard programming languages in ordinary digital computers, and are widely used in many fields of computer science. However, the ``indeterminacy'' is due to lack of knowledge or incomplete information, but not to actual indeterminate phenomena, that are not allowed in a fully deterministic view of the world. This observation may have little practical impact for current computing technology but it nevertheless leaves open the question of how the TM, the only model of computing, that is deterministic, accounts for truly non-determinate phenomena.

Genuine indeterminacy is measured with the entropy. Laplace's determinism can be considered as the limiting case in which every state of the world has only one possible next state. However, this condition can be relaxed allowing every state to have a number of possible next states, with the corresponding transition probabilities. Hence every state has an associated probability distribution of moving to the next state, and its indeterminacy is Shannon's entropy directly. The indeterminacy of the world is the average entropy of all the states. The next state of the world occurs randomly according to the probability distribution associated to the current state. However, the number of states of the world changes with entropic processes, which are not reversible. Laplace's determinism corresponds to a reversible world with zero entropy. Likewise, the standard formulation of the Turing Machine does not include the entropy because it is fully determinate and its entropy is zero too.

This limitation can be relaxed including the entropy in the formulation of the computing engine. A proposal to this effect is \emph{Relational-Indeterminate Computing} (RIC) \cite{Pineda-RC-libro-2021,pineda-eam-2021,morales-pineda-2022}. In this formalism the basic object of computing is the mathematical relation, instead of the function, and evaluating a relation is construed as selecting one of the objects associated to the given argument randomly. It is considered that this is a genuinely indeterminate choice, and a machine using this basic evaluation mechanism is indeterminate.

RIC has three basic operations: \emph{abstraction}, \emph{containment} and \emph{reduction}. Let the sets $A = \{a_1,...,a_n\}$ and $V = \{v_1,...,v_m\}$ be the domain and the codomain of a denumerable or a finite relation $r: A\to V$. For any relation $r$ there is a function $R: A\times V\to \{0,1\}$ such that $R(a_i,v_j) = 1$ or \emph{true} if the argument $a_i$ is related to the value $v_j$ in $r$, and $R(a_i,v_j) = 0$ or \emph{false} otherwise. Let $r_f$ and $r_a$ be two arbitrary relations from $A$ to $V$, and $f_a$ be a function with the same domain and codomain.  The operations are defined as follows:

\begin{itemize}
\item Abstraction: $\lambda(r_f, r_a) = q$, such that $Q(a_i, v_j) = R_f(a_i, v_j) \lor R_a(a_i,v_j)$ for all $a_i \in A$ and $v_j \in V$ --i.e., $\lambda(r_f, r_a) = r_f \cup r_a$.
\item Containment: $\eta(r_a, r_f)$ is true if $R_a(a_i,v_j) \to R_f(a_i,v_j)$ for all $a_i \in A$ and $v_j \in V$ (i.e., material implication), and false otherwise.
\item Reduction: $\beta(f_a, r_f) = f_v$ such that, if $\eta(f_a,r_f)$ holds $f_v(a_i) \in \{r_f(a_i)\}$ for all $a_i$, where  $f_v(a_i)$ is selected randomly, with an appropriate distribution centered around $f_a$. If $\eta(f_a,r_f)$ does not hold, $\beta(f_a, r_f)$ is undefined --i.e., $f_v(a_i)$ is undefined-- for all $a_i$.
\end{itemize}

The $\lambda$ and $\eta$ operations are fully determinate, but the relational evaluation $\beta$ makes a random choice and introduces the indeterminate character to this form of computing.

The form of RIC in which the objects of computing are finite relations represented as standard tables is called \emph{Table Computing}. The table's columns correspond to the relation's arguments, the rows to their values, and the cells are marked $1$ or $0$ depending on whether or not the corresponding argument and value are related. The $\lambda$ and $\eta$ operations are local to the corresponding cells of the tables representing the relations $f_a$ and $r_f$, and $\beta$ to the corresponding columns; and all three operations can be performed in parallel.

Relations have an associated entropy, which is defined as follows. Let $\mu_i$ be the number of values assigned to the argument $a_i$ in $r$; let $\nu_i$ = $1/\mu_i$ and $n$ the number of arguments in the domain. In case $r$ is partial, we define $\nu_i = 1$ for all $a_i$ not included in $r$. The \emph{computational entropy} $e(r)$ --or the entropy of a relation-- is defined here as: $$e(r) = -\frac{1}{n} \sum_{i=1}^{n} \log_2(\nu_i)$$

Functions are relations in which every argument has a unique value, or no value in the case of partial functions, but in any case are fully determined. Hence, the entropy of functions, either total or partial, is zero. The TM is the particular computing machine whose entropy is zero.

The number of functions $\xi$ included in a relation $r$ is the number of combinations that can be formed by selecting an argument of each column at a time, out of the $\mu_i$ possible values; if the relation is not defined for the argument $i$, $\mu_i=1$. This is: $$\xi(r)=\prod_{i=1}^{n}\mu_i$$ However, the entropy is the average indeterminacy of the arguments, and $\xi(r)=2^{e(r)^n}$ or simply $\xi(r)=2^{e(r)n}$. This can be seen directly considering that the indeterminacy of each argument is $-\log_2(1/\mu_i)$ = $-(\log_2(1)-log_2(\mu_i))=log_2(\mu_i)$, and $\mu_i=2^{log_2(\mu_i)}$; hence, $$\prod_{i=1}^{n}\mu_i=2^{e(r)n}$$.

Representations in RIC are construed as functions representing individual objects; and classes are construed as relations formed through the $\lambda$-abstraction of such functions. Individual membership is computed through the $\eta$ operation directly. In this representation the relation of a memory unit or unit of form --a marked cell-- may contribute to the representation of more than one unit of content or individual concept --a function-- which may share marked cells with other functions; and the relation between memory units and units of content is \emph{many-to-many}. The number of objects that are included in a relation is $\xi(r)$. This is a very large number, even for moderate values of the entropy and a relatively small number of arguments.

The basic RIC model has been enriched with a weighted version in which the content of the cells of the table storing the relation is an integer number representing the times such cell has been used in the representation of one or another function \cite{pineda-weam-2022,pineda-imagery-eam-2023}. In this version, the column corresponding to each argument becomes a probability distribution $\Psi$, whose entropy is Shannon's entropy directly; the entropy of the relation is the average entropy of all columns; and the argument of a $\beta$-reduction operation becomes a normal distribution on the column centered at the actual argument's value, named named $\zeta$. Let be $\Phi$ the scalar product of $\zeta$ and $\Psi$. The value $f_v(a_i)$ for each argument $a_i$ of the reduction operation $\beta(f_a, r_f) = f_v$, is selected randomly from $\Phi$. The number of functions stored in the relation $r$ is still $2^{e(r)n}$, although this is now the number of salient functions with a large enough weight, instead of the total number of functions.

Table computing can be thought of as a distributed and stochastic extension of the TM. The table plays the role of the tape, and the symbols stored in the cells are instances of the machine's alphabet. The machine has a scanner placed on top of each cell, allowing the parallel manipulation of cells and columns directly. The machine operations are stated in a transition table, but the algorithms implementing the $\lambda$, $\eta$ and $\beta$ operations are performed in very few computing steps, always terminate, and constitute \emph{minimal algorithms}. The conceptual extension is that representations are genuinely distributed and every state has an entropy, and the machine is intrinsically entropic.

\section{Entropic Associative Memory}
\label{EAM}

Table Computing has been used to build the Entopropic Associative Memory (EAM) \cite{pineda-eam-2021,morales-pineda-2022,pineda-weam-2022,pineda-imagery-eam-2023}. EAM uses one or more tables for representing objects, which are called \emph{Associative Memory Registers} (AMRs). Individual objects are represented through feature-value structures or functions, and memory register, recognition and retrieval are implemented with the $\lambda$, $\eta$ and  $\beta$ operations, respectively. The $\lambda$-register operations builds a distributed representation through the weighted disjunction of the registered objects; $\eta$-recognition rejects objects not contained in the memory using the logical material implication; and $\beta$-retrieval constructs the representation of a novel object randomly, out of the AMRs but modulated by the corresponding cue. The memory has been used to store, recognize and recover hand written digits, both with complete and partially occluded cues \citep{pineda-eam-2021}; manuscript letters and digits \citep{morales-pineda-2022}; store and learn phonetic information \citep{pineda-weam-2022}; and store, recognize and recover pieces of clothe, shoes and bags, with complete and noisy cues; to recover associated and imaged objects, and to make associations chains in which the object produced by a retrieval operation is used as the cue to a new retrieval operation \citep{pineda-morales-2022}. 

AMRs have an operational range of entropy values, and memory recognition and retrieval depend on the entropy of the AMR. For low entropy values precision is high but recall is low and vice versa; however, there is an entropy interval in which precision and recall are satisfactory; and the performance of the system as a whole obeys an entropy trade-off. If the entropy is very low, the retrieved objects are ``photographic'' reproductions of the cue, but recall is low; when the entropy is increased moderately, the retrieved objects are reasonable reconstructions of the cue or associated objects with a satisfactory compromise between precision and recall; at higher entropy values these become imagined objects, as recall is high but precision low; and if the entropy is very high, cues are always accepted but the recovered object may be just noise.

The Entropic Associative Memory resembles natural memory in that it is:
\begin{enumerate}
\item \emph{Associative}, in opposition to being accessed through addresses, such as standard RAMs used in symbolic systems;
\item \emph{Declarative}, as the $\lambda$-register, $\eta$-recognition and $\beta$-register operations are declarative, and the stored object are represented through overt functions, in opposition to sub-symbolic systems using numerical representations;
\item \emph{Distributed}, in opposition to local memories of TMs;
\item \emph{Parallel}, as memory register and memory recognition are cell-to-cell operations, and memory retrieval is a column-to-column operation, which can be implemented in parallel in very few computing steps if an appropriate hardware is provided;
\item \emph{Abstractive}, as the $\lambda$-register operation produces the disjunctive abstraction between the registered object and the content of the memory;
\item \emph{Productive}, as there are novel objects emerging from the combination of the objects stored explicitly; these objects do not constitute innate nor empirical knowledge, but are used as ``pivots'' for the recognition and construction of novel objects;
\item \emph{Strong negation}, as the recognition operation uses a declarative test --i.e., material implication-- in one computing step, in opposition to systems that ``negate'' by failing to find the sought object, which implement a form of the so-called close-world assumption;
\item \emph{Constructive}, as memory retrieval always produces a novel object, in opposition to RAM memories or Hopfield memories \cite{hopfield-1982} that always reproduce a previously registered object;
\item \emph{Indeterminate}, as the information is ``overlapped'' in the memory and the content is not transparent to direct inspection.
\end{enumerate}

A salient feature of the EAM is that the distributed property can be actual if the appropriate hardware is provided, and not only simulated, as is the case in computing paradigms using the TM. The distinction between local and distributed representations is reflected in the nature of the interpretation process: while the former is transparent and systematic --i.e., the meaning of the whole is a function of the meaning of the parts and their mode of grammatical composition, as in Frege's principle of compositionality-- the latter is performed through a holistic act that presents the meaning to the interpreter's mind directly, and cannot be analyzed. The properties of the EAM system implement such functionality directly, as the retrieved object is presented to the interpreter's eye as a whole in a single un-analyzable act.

The indeterminate property seems to be necessary not only to allow the storage of the huge number of objects that are registered in natural memories, but also to make them readily available when the cue ``impacts'' the indeterminate memory mass. This property can be better understood with an analogy to the interpretation of ambiguous images, such as Wittgenstein's famous duck-rabbit figure \citep{Wittgenstein-1953} or the visual images that emerge in diagrammatic proofs, such as the proof of the Theorem of Pythagoras \citep{pineda-2007}. The figures may not be assigned an interpretation at first sight, but suddenly an image appears in the interpreter's mind eye; although it may be unstable and switch spontaneously, and the interpreter may see another image. The phenomenon can be thought of as a recurrent memory process in which the object constructed by $\beta$-retrieval is used as the cue to a novel retrieval operation.

\section{The Mode of Computing}
\label{Mode-of-Computing}

\begin{figure}
\includegraphics[width=0.5\textwidth]{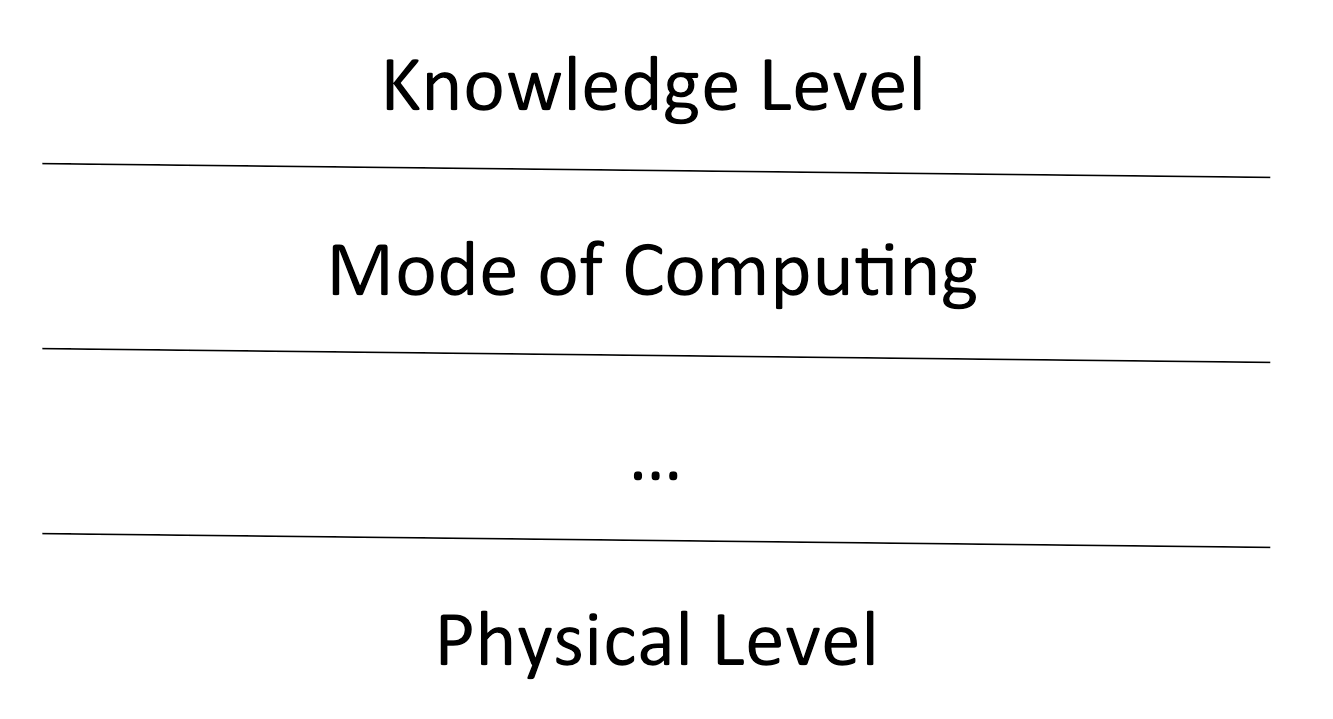}
\centering
\caption{The mode of Computing}
\label{Mode of Computing}
\end{figure}

The former discussion suggests that there is an open-ended set of intuitive notions of computing underlying different machines or formalisms. This diversity can be seen in terms of the proposed hierarchy of system levels in which there is a distinctive system level that here is called \emph{The Mode of Computing}. This level stands directly below the knowledge level and above other system levels down to the physical one, as illustrated in Figure \ref{Mode of Computing}. The mode of computing is a material process but the knowledge level consists on human interpretations.

The mode of computing is the physical device or artifact, either artificial or a natural phenomenon, that provides the material support and carries on with the operations that map the representation standing for the argument of a function or a relation into the representation standing for its value. In the case of the TM and its equivalent formalisms --the $\lambda$-\emph{calculus}, Recursive Functions Theory, \emph{Abacus} computations or the Von Neumann Architecture, ANNs, etc.--  the mode of computing is \emph{Algorithmic Computing}. This mode uses well-defined procedures or algorithms that map arguments into values of the function being computed by symbolic manipulation. Formal languages and automata theory, the theory of the complexity of algorithms, the theory of computability and non-computable functions, such as the halting problem, were developed in relation to algorithmic computing but other modes may have different salient features.

There are modes that use symbolic manipulation but differ from algorithmic computing in principled aspects. For instance, Relational-Indeterminate Computing is stochastic and includes the computational entropy. There are also modes of computing that do not use symbolic manipulation and perform computations by other means such as analogical and quantum computing, and even sensors and transducers of diverse sorts. These modes do not rely on algorithms: electrical analogical computers transform inputs into outputs almost instantly and there is no sense in which these machines compute following a well-defined discrete procedure. Quantum computers can also be thought of as analogical and the use of the term ``quantum algorithm" is informal for the same reason. Also, ANNs could be implemented with specialized hardware, using neuromorphic processors or other forms of unconventional computing, that do not use symbolic manipulation \citep{Ziegler2020}. 

The level of the mode of computing can be partitioned in two or more sub-levels; for instance, a program written in a procedural language, such as Fortran, Pascal or C, stands in a sub-level immediately above the same program but compiled in assembler, which in turns stands above of corresponding binary program.  Other examples are declarative programming languages such as Prolog or Lisp whose interpreters are written in C, which in turn are expressed in assembler and binary code. These levels constitute different levels of abstraction for human interpretation but convey a trade-off from top to bottom between expressiveness and efficient computation. The AI distinction between symbolic and sub-symbolic can be construed as sub-levels of the symbolic or algorithmic mode.

For analogical modes --such as quantum computing-- there may be a symbolic interface through which inputs and outputs to the natural phenomenon that performs the computing process --the actual quantum engine-- are presented to the knowledge level; similarly, standard analogical computers may have a symbolic interface through which the inputs and outputs are presented to the human interpreter.

%Different modes of computing may be integrated in complex systems to profit from their particular strengths. For instance, training current deep-learning ANNs machines can be thought of as learning a particular table for a very large domain beforehand, establishing a trade-off between strong algorithmic effort for learning and efficient computation for actual use. Similarly, current deep-reinforcement learning algorithms may be integrated with Relational-Indeterminate Computing to profit from both modes. For instance, the AlphaZero algorithm that recently beat the world champion programs in the games of chess, go and shogi, train a deep-learning convolutional network from the rules of the games and random play, using the Monte Carlo tree search algorithm, without additional sources of information \citep{silver-2018}. The input to the network is a board position $s$ and the output is a vector of move probabilities with components $p_a = Pr(a|s)$ where $p_a$ is the probability of wining the game if the action $a$ is made at position $s$. This strategy can be seen as building a huge relation from chess positions into moves, where each move has an associated probability.

However, there is no computation without the knowledge-level. The input and output of the computing engine at any mode of computing must be interpreted in relation to a predefined set of conventions, and the product of such interpretation is human knowledge. If there is computing there is interpretation but also if there is interpretation there is computing. A comprehensive notion of computing should involve both aspects of the phenomenon.

\section{The Mode of Natural Computing}
\label{mode-nat-comp}

Natural computing, in case it does exist, is carried on by natural brains of humans and animals with a developed enough nervous system. By analogy with the previous discussion there must be a system level directly below the knowledge level and above the physical brain structures that support the production of interpretations and experience that here is called \emph{The Mode of Natural Computing}. Such system level should specify the code in which information is expressed and its interpretation process, and there must be specific brain structures that support such functionality. The relation between regions of the brain and mental functions is very likely \emph{n-to-n} and the mode of natural computing may be supported by the functional organization networks, such as the orienting, alerting and executive attention networks, which play a strong role in the regulation of behavior, the control of affect and of the sensory input, that give rise to consciousness and voluntary behavior \citep{posner-2007, posner-2019}.

It is not necessary that all brain structures perform computations, which involve interpretations, and there may be structures that perform non-computational functions. In the world of artifacts invented by people there were control systems long before the TM was introduced and the notion of computing was available; similarly, in natural brains there may be structures performing control functions that precede the structures that are causal to consciousness and experience. Control structures are implemented by standard machinery that is there to be used but not to be interpreted, in opposition to computing machinery, whose only purpose is to map input into output representations at the mode of computing, that are interpreted and experienced by people.

The notion of the natural mode of computing is not necessary for non-representational and non-computational views of cognition, which deny that there is a computing engine that is causal and essential to the mind in the first place. For instance, views that the brain is a dynamical system from which the mind emerges, as discussed above, are non-computational. Computers may be used as modeling tools, that aim to describe the functionality of the brain through computational models, but the models themselves are neither causal and nor essential to the modeled phenomena; in the same sense that the computational models of the physical, chemical and biological world are not causal and essential to the actual physical, chemical and biological phenomena. Nature proceeds its course regardless it is modeled by a computer for the benefit of humans. Even computational models aiming to describe psychological phenomena are neither causal nor essential to the psychological world --unless such assumption is made explicitly, as in cognitive science and AI. A computational model requires representing the modeled object and make transformations upon such representation; but the representations are interpreted by people. In computing technology the mode of computing and the interpretation is performed by different entities, but in the case of the putative natural computing, the computing engine and the interpreter are the same individual. The whole idea that the mind is powered by a computing engine rests on the discovery and characterization of the mode of natural computing.

\section{Artificial versus Natural Computing}
\label{Art-Nat}

Artificial computing based on the TM supposes the intensive use of algorithms for representing knowledge, both conceptual and of abilities, and for performing inferences. Algorithms can be seen as intensional descriptions of the functions they compute, whose purpose is simply to make explicit their extensions. Computing proceeds serially, supported by a local scanner that inspects the memory --which consists of a ``passive recipient''-- according to the structural ingredients of the TM. Computations are deterministic and are performed with great precision and at a high speed. Computations are normally performed forwards --from arguments to values-- but not normally backwards --from values to arguments-- because not all functions have an inverse.

However, whether the mind uses algorithms --assuming that there is indeed natural computing-- is an open question. People are very limited to making extensive algorithmic calculations without the support of external aids and computing machinery. The information presented to living entities through perception is often extensional, and is plausible that a good deal of it is stored and processed in this form. Also, the amount of information and the structure of the brain suggest the use of highly distributed formats, with certain level of indeterminacy, that memory is fundamentally associative, and that computing is performed in parallel.

Natural computing, in addition, requires to perform inverse computations frequently; for instance, to recognize an object from its sensitive properties, or for diagnosing the causes of an event from its effects, often with uncertainty and indeterminacy. For this, it is plausible that natural computing uses formats that allow to establishing inverse associations, from values to arguments, very efficiently. ANNs have these properties to certain extent and reflect natural computing, but the need to simulate them in TMs limits significantly their explanatory power. Table and Relational-Indeterminate Computing, on their part, have such properties directly and may model better some aspects of natural computing.

\begin{figure}
\includegraphics[width=0.8\textwidth]{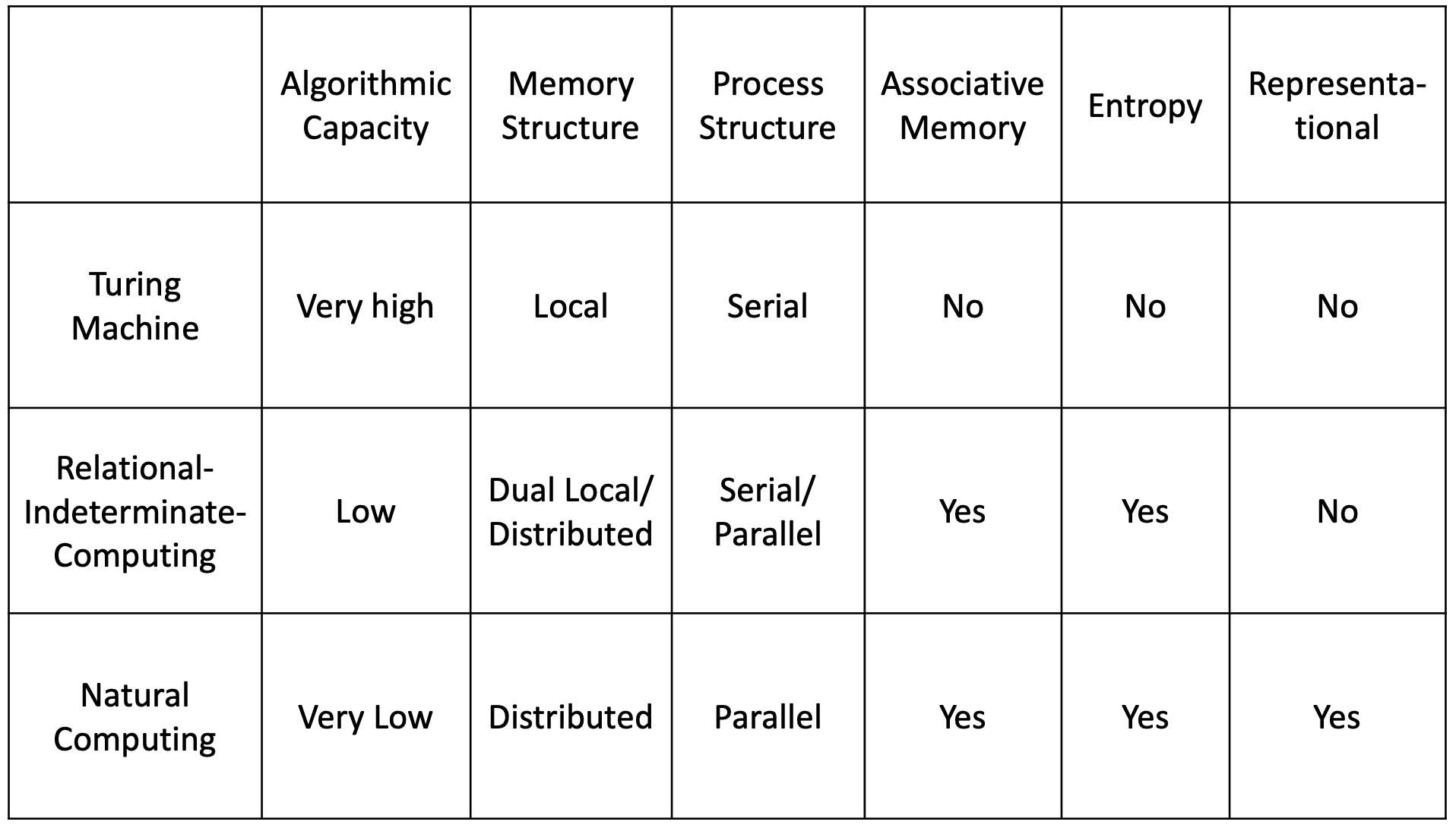}
\centering
\caption{Artificial versus Natural Computing}
\label{artificial-versus-natural}
\end{figure}

These considerations suggest to distinguish artificial or engineering computing, either in the form of the TM or RIC, versus natural computing in six salient dimensions: 1) algorithmic capacity; 2) memory structure; 3) whether the process is serial or parallel; 4) whether the mode supports associative memory; 5) whether the system is indeterminate and sustains a level of computational entropy; and 6) whether the computing agent sustains representations, and hence makes interpretations, experiences the world and can be conscious. The values of these dimensions for both artificial modes of computing and for natural computing are illustrated in Figure \ref{artificial-versus-natural}. It is also plausible that the brain uses simple algorithms but also other modes of computing, and that a trade-off between modes is established.

\section{Implications for Church Thesis}
\label{Church-Thesis}

A very strong current of opinion in computer science, and also of popular opinion, holds that the TM is the most powerful computing engine that can ever exist \citep{cop-church-turing}. The present discussion contests such position and it needs to be clarified the sense in which different modes of computing, including the mode of natural computing, differ from the standard theory of TM and computability theory.

Church thesis establishes that the set of functions that can be computed by the TM corresponds to the set of functions that can be computed intuitively by people, given enough time and paper. If a function does not have an algorithm it cannot be computed at all. If a non-computable function were computed by other means the Thesis would be refuted. Of course, if all functions were computable the Thesis would be empty. The thesis is based on the equivalence of all general enough formalisms or machines, such as the TM, the theory of recursive functions and the $\lambda$-Calculus, that compute exactly the same set of functions. The limits of computing depend on the impossibility to tell in advance whether a TM will halt for a given argument, the so-called Halting Problem. It is known that this problem cannot be solved by a TM. Hence, if a computing engine to such an effect were found --the so-called Halting-Machine-- it would not be a TM and, as the TM is the most powerful machine by hypothesis, the Halting Machine cannot exists at all. If such machine were ever discovered Church Thesis would be refuted too \cite{Boolos-Jeffrey}.

A first observation is that Church Thesis presupposes that the object of computing is the mathematical function; hence it does not apply to modes of computing that are based in alternative notions. Nevertheless, the extent of the thesis is explored next in some detail to clarify its scope and limitations. First I characterize the set of functions with finite domain and codomain, which are referred to here as \emph{finite functions}; and then discuss functions with infinite domains and codomains.

The set of all total and partial finite functions $F$ can be placed in a list. This set is different from the set of all total and partial functions, which is infinite and uncountable, as shown by Cantor's anti-diagonal arguments --see Boolos \& Jeffrey, 1989, cap. 2. 

Let the sets $A = \{a_1,...,a_n\}$ and $V = \{v_1,...,v_m\}$ with cardinalities $n$ and $m$ be the domain and codomain of the set of functions $F_{n,m}$ such that $n, m \geq 1$. The arguments and values represent arbitrary individual entities, either concrete or abstract, although abstracting from their extension and quality, and the only requirement is that they can be placed on a list. Every function $f_k \in F_{n,m}$ can be expressed in a table with $n$ columns and $m$ rows, such that every column has at most one marked cell, where columns represent the arguments and rows the values, and the marked cells represent the functional relation.

\begin{figure}
\includegraphics[width=0.2\textwidth]{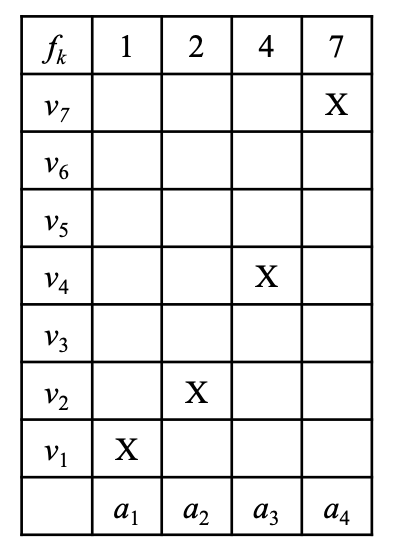}
\centering
\caption{Diagrammatic Representation of Finite Discrete Functions}
\label{diag-rep-func}
\end{figure}

Figure \ref{diag-rep-func} illustrates a table expressing a function whose domain and codomain are the sets  $A= \{a_1,a_2,a_3,a_4\}$ and $V = \{v_1,v_2,v_3,v_4,v_5,v_6,v_7\}$ with cardinalites $n=4$ and  $m=7$, respectively. The function has an index $f_k$ as shown in the top row, which in this case is $1247$. This is a number of $n$ digits in base  $m+1$ which is formed with the index  $j$ of the value $v_j$ for the argument $a_i$ in case the function is defined for such argument and with $0$ otherwise, for all the arguments. In particular, the index formed by a string of $n$ $0$s corresponds to the empty function that assigns no value to all of its arguments. The index as well as the names of the arguments and values are metadata, and are not considered part of the table, which is the object of computing proper.

The figure illustrates that the value of $f_k$ ranges from ``$0000$'' to ``$7777$'' in base $8$ --i.e., from $0$ to $4095$ in base $10$. In general, the number of a function is $N_k = (f_k)_{10} + 1$, where $(f_k)$ is a number in base  $m+1$ and  $(f_k)_{10}$ denotes $f_k$ in base $10$. In the example,  $N_k = (f_k)_{10} + 1=1247_{10}+1= 680$.

The number of total and partial functions that can be formed with $n$ arguments and $m$ values is $(m+1)^n$. This can be seen directly by noticing that the number of function is simply the number of combinations that can be formed with $n$ arguments, and each can be associated to $(m+1)$ values. The number of functions in Figure \ref{diag-rep-func} is $(7+1)^4 = 4096$.

Next, we order the set of functions $F_{n,m}$ that can be included in a table of size $n \times m$ for $m, n \geq 1$. All order pairs $(n,m)$ can be placed in a list; consequently, all sets $F_{n,m}$ can be placed in a list. Let $t$ be the index of a set $F_{n,m}$.  Figure \ref{order-grids} shows a diagonal order of all tables of size $n \times m$ for $m, n \geq 1$, where each table is labeled with its index $t$. In particular,  $t=1$ is the index of the functions $F_{1,1}$ that have one argument and one value at the most. This set has two functions: $f_0$ that assigns no value to its argument, and $f_1$ that does so. In order to assign the index $t$ to an arbitrary finite function as a function of $n$ and $m$:

\begin{figure}
\includegraphics[width=0.4\textwidth]{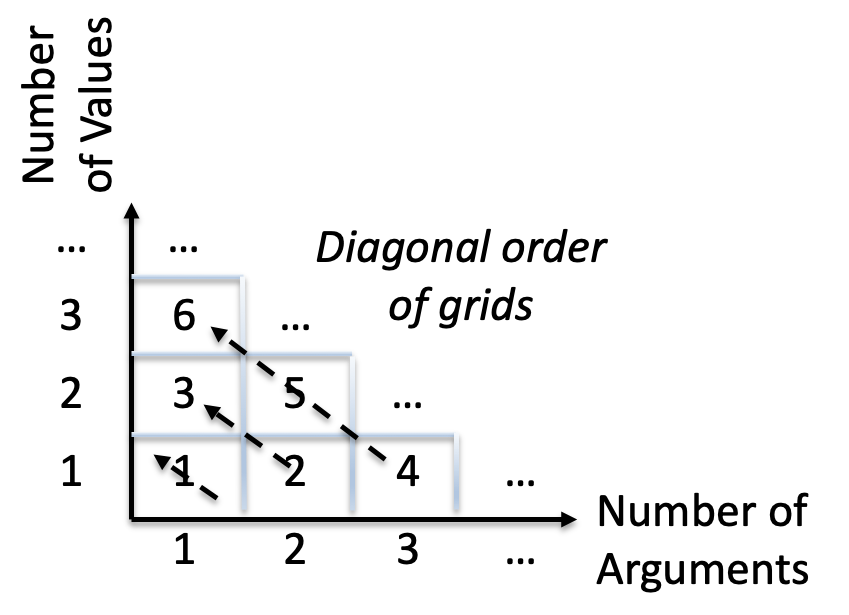}
\centering
\caption{Order of Tables for Finite Discrete Functions}
\label{order-grids}
\end{figure}

\begin{enumerate}
\item It can be seen by direct inspection of Figure \ref{order-grids} that:
\begin{itemize}
\item Each diagonal $j$ has $j$ tables;
\item The index $t$ of each table is increased over the diagonal from bottom to top and from right to left;
\item The table $n \times m$ lays on the diagonal $diag(n,m) = n + m -1$.
\end{itemize}
\item It can be seen by diagrammatic reasoning that:
\begin{itemize}
\item The largest index in diagonal $j$ is defined recursively as follows: $max(0) = 0$ and $max(j) = max(j-1)+j$;
\item $t$ = $max(diag(n,m)-1)+m$.
\end{itemize}
\end{enumerate}

The number of a function $N_f$ in table $t$ is the sum of all the functions in tables $1$ to $t-1$ added on the function $N_k$ in the table $t$. Hence, for finding out the number of a function in the table with cardinality ($n$,$m$) it is required to find the index $t$ as a function of $n$ and $m$. This is as follows: 

\begin{enumerate}
\item Let $diag(i)$ the diagonal of table $i = (n(i),m(i))$;
\item It can be seen through diagrammatic reasoning that:
\begin{itemize}
\item The function $diag(i)$ is computed through the following minimization: $diag(0) = 0$ and $diag(i) = j$ such that $j=0$; increment $j$ until $max(j-1) < i \leq max(j)$;
\item $n(i) = max(diag(i)) - i +1$;
\item $m(i) =  i - max(diag(i)-1)$.
\end{itemize}
\end{enumerate}

The absolute number of function $N_f$ is the function $f_k$ whose domain and codomain with cardinality $n$ and $m$ is:

 \begin{mydef}
{\bf Number of function for $n, m \geq 1$} \\
\vspace{0.4cm}
\hspace{1.5cm} 
$N_f = N_k$ if $t=1$; \\
\vspace{0.1cm}
\hspace{1.5cm} 
$N_f= \sum_{i=1}^{t-1}(m(i)+1)^{n(i)} + N_k$ if $t >1$
\label{order-funcs-formal}
\end{mydef}

Definition \ref{order-funcs-formal} is recursive and can be computed by a TM. It establishes that all total and partial functions with finite domain and range are enumerable, regardless the size of $n$ and $m$. So, given the size of the table and a function in such table, the absolute number of such function is provided. Conversely, all functions can be generated out of their indices. The procedure is simply iterate over the index $i$ from $i = 1$ to $t$; compute $n(i)$ and $m(i)$; and compute all indices of length $n(i)$ in base $m(i)+1$ --i.e., from $0...0$ to $m...m$. This renders all tables or the extensional representation of all finite functions.

This enumeration is redundant. In particular all set of function $F_{n,m_j}$ includes all functions in the set $F_{n,m_i}$ if $j > i$. Consequently, all functions have a enumerable set of indices, but the enumeration is complete, and picks up all finite functions with cardinalities $n, m \geq 1$.

All finite functions in the table format can be computed by one single extensional table-computing non-recursive procedure, which consists of selecting the column corresponding to the argument, finding out the marked cell in such column, and selecting the value associated to the row. Table-computing can be thought of as a TM that uses the table as its tape, and the table-computing algorithm is recursive, although trivially. Alternatively, all finite functions can be computed extensionally by a TM by placing all the pairs of arguments and values on the tape, in the order given by Definition \ref{order-funcs-formal}, and scanning the tape until the given argument is found and its associated value is returned. It follows that all finite functions are recursive, but trivially too. Hence, there are none non-computable finite functions, and Church Thesis holds but is empty for this set --i.e., the set of recursive finite functions and the set all finite functions is the same. Furthermore, such extensional machines provide the values of all the arguments of all finite functions by direct inspection and are immune to the halting problem.

However, algorithmic computing is generally understood as computing functions through effective procedures. Standard programs running in ordinary digital computers, that have a finite-size processing and memory registers (e.g., 32 or 64 bits), and finite memory, are the paradigmatic case of such non-trivial algorithms that compute finite functions. Computable functions have an underlying structure or regularity that is capitalized in the construction of their algorithms, but the table format suggests that there is a large number of finite functions that have none or very little structure, that may not have a non-trivial algorithm and would not be computable. If at least one such function were found, it could nevertheless be computed through table-computing, and Church Thesis would not hold in this non-trivial sense. The enumeration also shows that the halting problem for non-trivial algorithms computing finite functions could be solved if the number of a function can be determined out of the specification of the algorithm and vice versa, although it is very unlikely that such correspondence can ever be established. 

For these reason the table and the algorithmic are different modes of computing. Table-computing can be employed only when the function is provided extensionally, which is plausible in natural settings and concrete problem-solving, but the extension of most mathematical functions is known intensionally and non-trivial algorithms are needed to compute them. Humans are very limited for computing algorithms and that is why computers were invented in the first place.

Now we turn to the case in which the domain and range of the functions are not finite. The TM is an abstract machine defined to have a enumerable infinite set of inputs. Hence, in addition to the finite functions enumerated by Definition \ref{order-funcs-formal}, there are uncountable many functions with infinite domains and codomains. One first intuition to characterize such functions is that, as the parameters $n$ and $m$ are infinitely many, these can be extended \emph{ad infinitum}. However, finite and infinite sets are different kinds of objects and such identification would be a category mistake. In the infinite case, there may be a very large number of functions that lack an algorithm and are not recursive, which of course would not have a table representation, and could not be computed through other means. These are the non-computable functions. If a single functions of this kind could be computed by other means, Church Thesis would be refuted, in the intended relevant sense.

Computational models of physical systems are stated through equations with real domains and codomains, and their computation requires algorithms with infinite many inputs. However, real numbers can be approximated accurately enough through their discrete representation in digital memory registers, and physical systems can be modeled through digital computers with the only limitation of the available time and memory resources. It may also be the case that the solution of a model of a physical system is a non-computable function, in which case it cannot be computed neither by people nor by machines. If a model of a physical system whose solution is a non-computable function were solved by other means, Church Thesis would be refuted too.

The discussion up to this point shows the sense in which the TM is the most powerful computing machine, which can be summarized as stating that there is an effective procedure or algorithm for each computable function, with the only limitation established by the halting problem; and that there are functions that cannot be computed in an absolute sense. Functional evaluation is a fully determined process by necessity. Hence, the indeterminacy of stochastic processes and non-deterministic automata is dissolved in a theory of functions.

Next, we address the sense in which the TM can be the most  ``powerful machine''. This is a complicated notion in several respects, starting with the fact that different machines or formalisms can be more powerful or weaker than others in particular aspects, even though all of them may compute the full set of computable functions. For instance the TM is better than the von Neumann architecture to study the general properties of computing machines, but the von Neumann architecture is better to implement practical computations, despite that both of these formalisms compute the same set of functions. Also, all models of computing assume particular trade-offs; for instance, algorithmic computing assumes a fundamental trade-off between the expressive power of representations and their tractability, or the possibility of perform a computation with finite time and memory resources. This trade-off is explicit in Chomsky's hierarchy of formal languages that includes, from bottom to top, the regular languages, the context free languages, the context sensitive languages and the languages without restrictions of any sort \citep{Hopcroft}. Regular languages have a concrete character and can be computed very efficiently but can express only very limited abstractions while the languages in the other extreme can express very deep abstractions but the computational cost can be very high and cannot be afforded always. The algorithmic mode also assumes the \emph{The Knowledge Representation Trade-Off} that states that concrete representations can be computed efficiently but have expressiveness restrictions and that abstract representations can very expressive but cannot be computed effectively \citep{Levesque-brachman, Levesque}. For instance, the consequences of the knowledge expressed in propositional logic can be found very easily but if the full expressive power of predicate logic is required the computational cost may be too high and computations may be unfeasible. Another example is ANNs that can simulate distributed representations very efficiently but cannot express syntactic structure and cannot store declarative information.

Similarly, there are symbolic modes more powerful than a TM in particular respects. For instance, while the TM is directed to make complex calculations but is not decidable due to the halting problem, computing with tables is oriented to make direct associations and computations always terminate. Also, while computing a function and its inverse, whenever it exists, requires different algorithms, computing with tables provides the inverse function or a relation directly by inspection using the same algorithm for all inverse functions and relations. Another example is the opposition between the TM and Relational-Indeterminate Computing: while the former is deterministic the latter is indeterminate, stochastic and entropic. From this perspective the TM is the machine with zero entropy, but there are computing engines that have an entropy larger than zero. On its part, analogical computers are very efficient and computations are performed through the physical phenomena almost instantly but compute specific set of functions, are indeterminate and have no memory capabilities. 

More generally, every mode of computing assumes some fundamental trade-offs that define the its explanatory capability and its potential applications. There may be properties of different modes that can be compared directly, such as the speed or memory capacity of digital versus quantum computers, but comparing different modes of computing in general is a category mistake.

The view that the mind is powered by a computing engine --hence there is natural computing-- also called ``computationalism'', holds that the TM computes all functions that can be computed intuitively by people, and assuming the Church Thesis is true, mental processes are computational. However, this proposition presupposes that computing is evaluating a function independently of its interpretation, which is performed by people; or, alternatively, that computing is an objective property of mechanisms, but independent of the interpretation process. In the refutation of so-called ``Systems Reply'' of Chinese room story --the counter argument that consciousness is a property of the system as a whole but not only of the person in the room-- Searle argues convincingly that if such refutation were sound, which is not, it would also follow that every device, natural or artificial, that transforms inputs into outputs, such as biological organs --e.g., stomachs, livers and harts-- and also artifacts --e.g., thermostats, telephones, adding machines, electric switches, etc.-- should be granted consciousness too \citep{Searle}. In any case, sustaining that the mind is a computational process on the supposition that Church Thesis is true does not follow even within functionalism \citep{piccinini-2007}.

Despite all this, the notion of effective procedure or Turing's computability has been extrapolated and Church-Turing Thesis has been interpreted as stating that the TM can simulate all possible mechanisms, in particular in the forms that Copeland  \citep{cop-church-turing} calls the Maximality Thesis, sustaining that all functions that can be generated or performed by machines can be computed by a TM; the Simulation Thesis, holding that Turing's results imply that the brain, and any physical or biological system can be simulated by a TM; and, more radically, the Physical Church-Turning Thesis stating that any physical computing device or any physical thought experiment that is conceived or designed by any future civilization can be simulated by a TM. These theses taken together and other similar propositions are informally called the strong version of Church Thesis stating that the TM is the most powerful computing machine that can ever exist in any possible sense. These views seem to presuppose that the universe or nature is governed by a set of causal fundamental rules, that determine all possible events, at all system levels, that can be characterized as functions, in the mathematical sense, and that the TM is the deterministic device that can compute such functions. This is a reissue of Laplace's determinism, that is hard to die.

\section{Computing, Consciousness and Experience}
\label{Cog-Mode}

The computational metaphor of mind rests on the discovery and characterization of the mode of natural computing. Without such construct computing has no explanatory power in cognition and related disciplines. Artificial Intelligence and cognition are traditionally thought of within some of another form of functionalism that is realized by the TM, giving rise to the so-called functional computationalism \citep{piccinini-2010}, but this view is limited because the TM cannot make interpretations, and consciousness is not explained. Mental states of humans and no human animals with a developed enough nervous system have an ``intrinsic intentionality'' --e.g., \cite{egan-2014}-- but the states of the TM do not have such property, neither in the structural or functional ingredients of its abstract specification nor in its physical realizations. Computing states are material configurations, such as the gears and levers of Babbage's analytical engine, that needed to be interpreted by people as computing states. The assumption that embedding a physical realization of a TM --or of another artificial mode of computing-- in a physical body that interacts with the environment, provides intentionality to the machine states, due to such sole fact, does not follow, as discussed by Searle in his refutation to the System Reply \cite{Searle}. Closing the gap between computing states and mental or intentional states seems to be an utopian enterprise, even for the original proponents of functionalism \cite{Shagrir-2005}. Nevertheless, supporters of such doctrine claim that most criticism to functional computationalism are miss-understandings of one or another sort, including Searle's arguments, due to his position that consciousness is a subjective phenomenon, that is realized in biological brains, although it is not known why \cite{milkowski-2018}. The confusion comes from the identification of the notions of artificial and natural computing, which seems to be taken as the same by the critics, but it is clear that Searle is talking about the limitations of the TM, not of natural minds.

The assumption that the mode of natural computing does exist, in conjunction with the hierarchy of system levels, provides a novel perspective on the declarative versus procedural representations. In the former, the knowledge level holds representations of interpretations of the objects or processes at the level of the mode of natural computing, but in the latter such mental objects do not exist.

The ontological status of mental objects at the knowledge level is problematic, but if these objects are postulated at all, a parsimonious criteria about their nature is considering that they reflect the properties or processes of their corresponding objects at the level of the mode of computing, for all modes that are used in natural computing. This view is called here \emph{the representational view}. The algorithmic mode uses symbolic manipulation whose interpretations at the knowledge level can be thought of as propositions. Fodor's mentalese corresponds to this view. Other symbolic modes, such as Relational-Indeterminate Computing, are characterized by the use of the space as the medium and the indeterminacy of its structures. Representations at the knowledge level would require no further interpretation; otherwise, interpretation would involve an infinite regress. In the representational view it has to be held as well that human interpreters are conscious by the mere fact of having such representations in mind, at least when they are are the focus of attention, including the qualitative, experiential or phenomenological aspect of consciousness.

Procedural representations, on the other hand, do not involve that there are mental objects at the knowledge level. This can be elicited as claiming either that the so-called knowledge level consists of a continuous interpretation process of the objects at the level of the mode of natural computing, which we call \emph{the interpretative view}; or more radically, that the knowledge level does not exist at all, neither in artificial nor in natural computing, which we call \emph{the non-interpretative view}.

The interpretative view allows for holding that in natural computing the knowledge level and the mode of computing fuse in a single level that is causal and essential to consciousness. If the mode of natural computing uses a representational structure, such structure would hold the actual code in which information is stored in the brain, and the operations on it are actual interpretations that allow the agent to be aware, understand and experience the world. Such representations would be stable and could be consciously inspected in distinctive brain states. If there are not stable representational structures at the mode of natural computing, the interpretation is performed upon to the input and output information of continuous processes. The interpretation is also causal to experience and consciousness, but the states are too proximal between each other, as the individual pictures of a movie, that produce a continuous experience, but are not subject to conscious inspection. The interpretative is perhaps more parsimonious than the representational view, which can be dispensed with altogether, and allows for declarative and procedural natural modes of computing, that can co-exist coherently. 

Lastly, in the \emph{the non-interpretative view} the knowledge level is absent altogether in both artificial and natural computing; there are not interpretations, neither meanings, nor consciousness nor experience. In this perspective, computing is an objective property of mechanisms which can be ascribed to any kind of artifact and even to organs, as pointed out by Searle. Machinery would carry on computing even if the whole of the human kind --as well as all no-human animals endowed with natural computing-- were wiped out from the face of the earth. Furthermore, arbitrary physical, chemical or biological phenomena could also be considered computing machines, including the universe as a whole. But then, the whole notion of computing becomes rather empty.

However, there is no computation without representation and interpretation. Computing is not an objective property of mechanisms, but a relational phenomenon, involving a mode of computing and a conscious interpreter. The interpretation is causal to experience, to the ascription of meaning and to consciousness. Experience or awareness would be how computing with a particular mode feels, for all modes used in natural computing. Computing, interpreting and consciousness are three aspects of the same phenomenon; or consciousness and experience are the manifestations of computing/interpreting. Interpreting is what distinguishes computing machinery from standard machinery. In the case of artificial computing the process is split in two different entities: the one that supports a mode of computing and the one who makes the interpretation. In the case of natural computing the entity that holds the mode of computing and the one that makes the interpretation is the same individual.

The present proposal does not advocates rejecting that the mind is a computational processes, but holds that finding the mode or natural computing is required. This is an open question for science altogether and there is not or does not seem to be any plausible answer in sight. But for the time being, in the absence of the characterization of the mode or modes of natural computing, computing has no explanatory power in cognition and related disciplines, and can be removed altogether from current talk, such as other theoretical constructs that were held as fundamental for some time but lacked a real explanatory power and were dropped on the face of the evidence, such as the ether. If it turns out that the mode of natural computing does not exist, the mind would not be a computational process, but something else; conversely, discovering it would come with finding the property that makes interpreters ascribe meanings to symbols or processes and have subjective experience, and would equate with solving the hard problem of consciousness.

%\section{Acknowledgments}

\nolinenumbers

%\section{Bibliography styles}
%There are various bibliography styles available. You can select the 
%style of your choice in the preamble of this document. These styles are 
%Elsevier styles based on standard styles like Harvard and Vancouver. 
%Please use Bib\TeX\ to generate your bibliography and include DOIs 
%whenever available.
%Here are two sample references: \citep{Feynman1963118,Dirac1953888}.

%\section{References}

%\bibliographystyle{elsarticle-num}
\bibliographystyle{model-5names}
\bibliographystyle{apacite}

\bibliography{references}

\end{document}